\documentclass[table]{article}
\pdfoutput=1
\usepackage{iclr2017_conference,times}
\usepackage[utf8]{inputenc} 
\usepackage[T1]{fontenc}    
\usepackage{hyperref}       
\usepackage{url}            
\usepackage{booktabs}       
\usepackage{amsfonts}       
\usepackage{nicefrac}       
\usepackage{microtype}      
\usepackage{graphicx}
\usepackage{natbib}
\usepackage{xcolor}

\newcommand{\Co}{Concierge}
\newcommand{\mcb}[1]{\multicolumn{2}{c|}{#1}}
\newcommand{\mc}[1]{\multicolumn{2}{c}{#1}}
\definecolor{dgreen}{rgb}{0.0,0.4,0.0}
\definecolor{dred}{rgb}{1,0.0,0.0}
\usepackage{amsmath} 
\usepackage{amssymb}

\title{Learning End-to-End Goal-Oriented Dialog}

\author{Antoine Bordes, Y-Lan Boureau \& Jason Weston \\
Facebook AI Research\\
New York, USA\\
\texttt{\{abordes, ylan, jase\}@fb.com}
}

\iclrfinalcopy 

\begin{document}

\maketitle

\begin{abstract}
  Traditional dialog systems used in goal-oriented applications
  require a lot of domain-specific handcrafting, which hinders scaling up to new
  domains.  End-to-end dialog systems, in which all components are trained from
  the dialogs themselves, escape this limitation.  But the
  encouraging success recently obtained in chit-chat dialog may not carry over to
  goal-oriented settings.  This paper proposes a testbed to break down
  the strengths and shortcomings of end-to-end dialog systems in goal-oriented
  applications.
  Set in the context of restaurant reservation, our tasks require
  manipulating sentences and symbols in order to properly conduct
  conversations, issue API calls and use the outputs of such calls.
  We show that an end-to-end dialog system based on Memory Networks
  can reach promising, yet imperfect, performance and learn to perform
  non-trivial operations. We confirm those results by comparing our
  system to a hand-crafted slot-filling baseline on data from the
  second Dialog State Tracking Challenge \citep{henderson2014second}.
  We show similar result patterns on data extracted from
  an online concierge service.
\end{abstract}

\section{Introduction}
\vspace{-1ex}

\begin{figure}[ht]
  \begin{center}
    \includegraphics[width=0.8\linewidth]{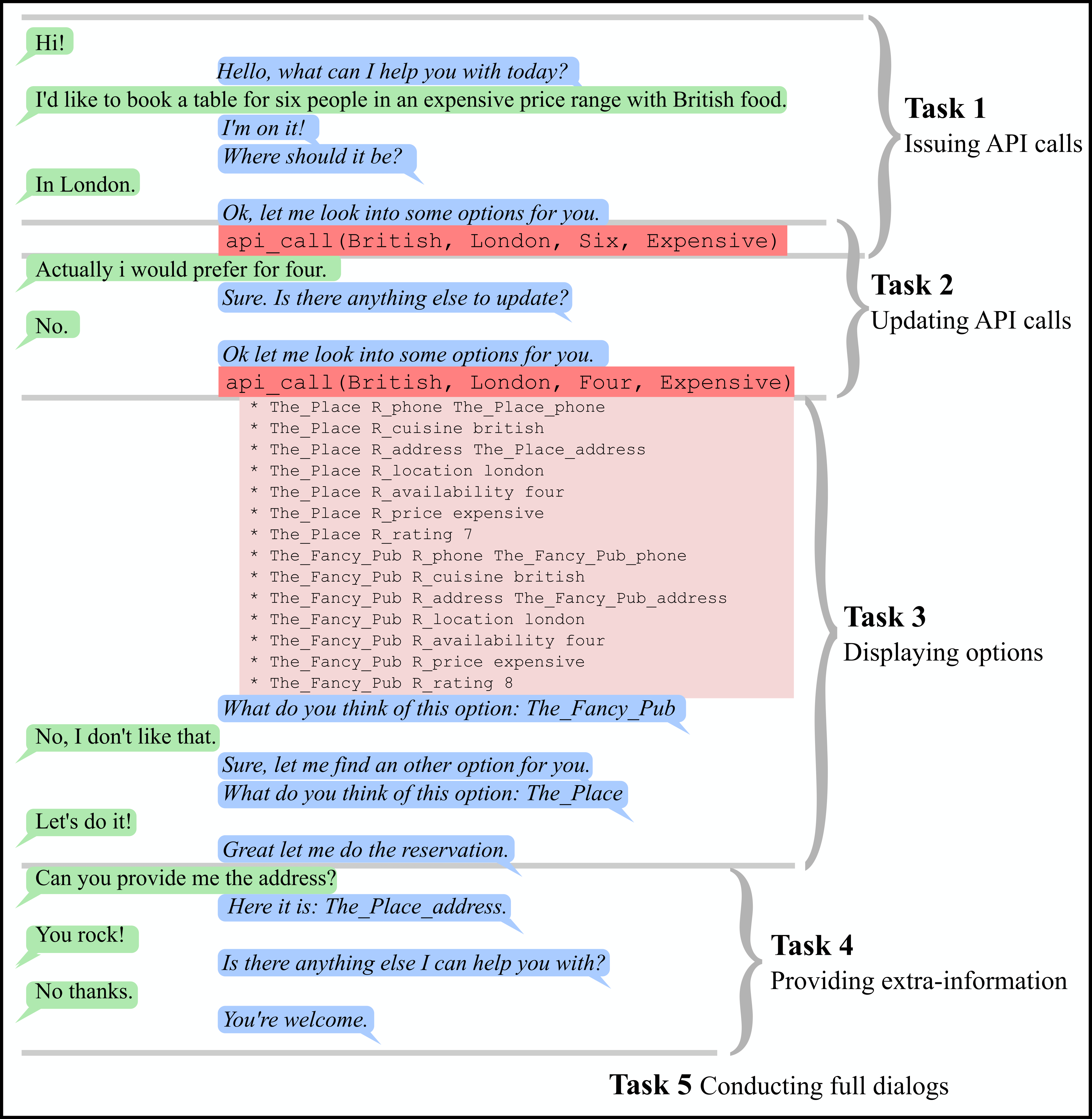}
  \end{center}
  \caption{\label{fig:tasks}{\bf Goal-oriented dialog tasks.} {\small A user
    (in green) chats with a bot (in blue) to book a table at a
    restaurant. Models must predict bot utterances and API calls (in
    dark red).  Task 1 tests the capacity of interpreting a request
    and asking the right questions to issue an API call. Task 2 checks
    the ability to modify an API call. Task 3 and 4 test the capacity
    of using outputs from an API call (in light red) to propose
    options (sorted by rating) and to provide extra-information. Task
    5 combines everything.}}
  \vspace{-2ex}
\end{figure}


The most useful applications of dialog systems such as digital personal
assistants or bots are currently goal-oriented and transactional: the system
needs to understand a user request and complete a related task with a clear
goal within a limited number of dialog turns. 
The workhorse of traditional dialog systems is
\emph{slot-filling} \citep{lemon2006isu,wang2013simple,young2013pomdp} which predefines the
structure of a dialog state as a set of slots to be filled during the dialog. 
For a restaurant reservation system, such slots can be the location, price
range or type of cuisine of a restaurant.
Slot-filling has proven reliable but is inherently hard to scale to new
domains: it is impossible to manually encode all features and slots that users
might refer to in a conversation.

End-to-end dialog systems, usually based on neural networks
\citep{shang2015neural,vinyals2015neural,sordoni2015neural,serban2015building,dodge2015evaluating},
escape such limitations: all their components are directly trained on past
dialogs, with no assumption on the domain or dialog state structure, thus
making it easy to automatically scale up to new domains.
They have shown promising performance in non goal-oriented
\emph{chit-chat} settings, where they were trained to predict the next
utterance in social media and forum threads
\citep{ritter2011data,wang2013dataset,lowe2015ubuntu} or movie
conversations \citep{banchs2012movie}.
But the performance achieved on chit-chat
may not necessarily carry over to goal-oriented conversations. As illustrated
in Figure~\ref{fig:tasks} in a restaurant reservation scenario, conducting
goal-oriented dialog requires skills that go beyond language modeling, e.g.,
asking questions to clearly define a user request, querying Knowledge Bases
(KBs), interpreting results from queries to display options to users or
completing a transaction. This makes it hard to ascertain how well end-to-end
dialog models would do, especially since evaluating chit-chat performance
in itself is not straightforward \citep{liu2016not}.
In particular, it is unclear if end-to-end models are in a position to
replace traditional dialog methods in a goal-directed setting: can end-to-end
dialog models be competitive with traditional methods even in the well-defined
narrow-domain tasks where they excel? If not, where do they fall short?

This paper aims to make it easier to address these questions by proposing an
open resource to test end-to-end dialog systems in a way that 1) favors
reproducibility and comparisons, and 2) is lightweight and easy to use.  We aim
to break down a goal-directed objective into several subtasks to test some
crucial capabilities that dialog systems should have (and hence provide error
analysis by design). 
In the spirit of the bAbI tasks conceived as question
answering testbeds \citep{weston2015towards}, we designed a set of
five tasks  within the goal-oriented context of restaurant
reservation. Grounded with an underlying KB of restaurants and
their properties (location, type of cuisine, etc.), these tasks
cover several dialog stages and test if models can learn various
abilities such as performing dialog management, querying KBs,
interpreting the output of such queries to continue the conversation or
dealing with new entities  not appearing in dialogs from the
training set.
In addition to showing how the set of tasks we propose can be used to test
the goal-directed capabilities of an end-to-end dialog system, we also
propose results on two additional datasets extracted from real interactions
with users, to confirm that the pattern of results observed in our tasks
is indeed a good proxy for what would be observed on real data, with the
added benefit of better reproducibility and interpretability.

The goal here is explicitly not to improve the state of the art in the narrow
domain of restaurant booking, but to take a narrow domain where traditional
handcrafted dialog systems are known to perform well, and use that to 
gauge the strengths and weaknesses of current end-to-end systems with
no domain knowledge.
Solving our tasks requires manipulating both natural language and symbols
from a KB.
Evaluation uses two metrics, per-response and per-dialog
accuracies, the latter tracking completion of the
actual goal. Figure~\ref{fig:tasks} depicts the tasks and
Section~\ref{sec:tasks} details them.
Section \ref{sec:models} compares multiple methods on these tasks.
As an end-to-end neural model, 
we tested Memory Networks \citep{weston2014memory}, an
attention-based  architecture that has proven
competitive for non goal-oriented dialog \citep{dodge2015evaluating}.
Our experiments in Section~\ref{sec:exp} show that Memory Networks can
be trained to perform non-trivial operations such as issuing API calls to KBs
and manipulating entities unseen in training.
We confirm our findings on real human-machine dialogs from the
restaurant reservation dataset of the  2$^{nd}$ Dialog State Tracking
Challenge, or DSTC2 \citep{henderson2014second}, which we converted into our task
format, showing that Memory Networks can outperform a
dedicated slot-filling rule-based baseline.
We also evaluate on a dataset of human-human dialogs
extracted from an online concierge service that books
restaurants for users.
Overall, the per-response performance is encouraging, but the 
per-dialog one remains low, indicating that end-to-end models still
need to improve before being able to reliably handle
goal-oriented dialog.

\vspace{-0.25ex}
\section{Related Work}
\vspace{-0.5ex}
The most successful goal-oriented dialog systems model conversation as
partially observable Markov decision
processes (POMDP) \citep{young2013pomdp}. However, despite recent efforts
to learn modules \citep{henderson2014word}, they still require many
hand-crafted features for the state and action space representations,
which restrict their usage to narrow domains.
%
Our simulation, used to generate goal-oriented datasets, can be
seen as an equivalent of the user simulators used to train
POMDP \citep{young2013pomdp,pietquin2013survey}, but for training
end-to-end systems.
%

\cite{serban2015survey} list available corpora for training dialog systems.
Unfortunately, no good resources exist to train and
test end-to-end models in goal-oriented scenarios.
Goal-oriented datasets are usually designed to train or
test dialog state tracker components \citep{henderson2014second}
and are hence of limited scale and not suitable for end-to-end
learning (annotated at the state level and noisy). 
However, we do convert the  Dialog State Tracking Challenge data into our framework.
Some datasets are not open source, and require a particular license agreement
or the participation to a challenge (e.g., the end-to-end task of DSTC4 \citep{kim2016fourth}) or are
proprietary (e.g., \cite{chen2016end}). Datasets are often based on
interactions between users and existing systems (or ensemble of systems) like
DSTC datasets, SFCore \citep{gavsic2014incremental} or ATIS \citep{dahl1994expanding}. This creates noise
and makes it harder to interpret the errors of a model. Lastly, resources
designed to connect dialog systems to users, in particular in the context of
reinforcement learning, are usually built around a crowdsourcing setting such
as Amazon Mechanical Turk, e.g.,
\citep{hixon2015learning,wen2015semantically,su2015learning,su2015reward}.
While this has clear advantages, it prevents reproducibility and consistent
comparisons of methods in the exact same setting.

The closest resource to ours might be the set of tasks described
in \citep{dodge2015evaluating}, since some of them can be seen as
goal-oriented. However, those are  question answering tasks rather than
dialog, i.e. the bot only responds with answers,
never questions, which does not reflect full conversation. 

\section{Goal-Oriented Dialog Tasks} \label{sec:tasks}
\vspace{-0.5ex}

All our tasks involve a restaurant reservation system,
where the goal is to
book a table at a restaurant. 
The first five tasks are generated by a simulation,
the last one uses real human-bot dialogs.
%
The data for all tasks is available at \url{http://fb.ai/babi}.
We also give results on a proprietary dataset
extracted from an online restaurant
reservation concierge service with anonymized users.

\begin{table}[t]
  \newcommand{\mcf}[1]{\multicolumn{5}{c|}{#1}}
  \newcommand{\mco}[1]{\multicolumn{1}{c|}{#1}}
  \newcommand{\mcr}[1]{\multicolumn{2}{r|}{#1}}
  \begin{center}
        \caption{\label{tab:data}{\bf Data used in this paper.} {\small Tasks
          1-5 were generated using our simulator and share the same
          KB. Task 6 was converted from the 2$^{nd}$ Dialog State
          Tracking Challenge \citep{henderson2014second}.
	  \emph{\Co} is made of chats extracted from a real
          online concierge service.
          $^{(*)}$
          Tasks 1-5 have two test sets, one using the vocabulary of
          the training set and the other using out-of-vocabulary
          words. }}
      {\small
    \begin{tabular}{c|l|ccccc|c||c}
       \mcr{\bf Tasks} & \bf T1& \bf T2& \bf T3 & \bf T4 & \bf T5& \bf T6& \bf \Co  \\
      \hline
              & Number of utterances:             & 12 & 17 & 43 & 15 & 55 & 54 & 8\\
      {\sc Dialogs} & \mbox{~~~~}- user utterances        &  5 &  7 &  7 &  4 & 13 &  6 & 4 \\
       \mco{{\small {\it Average statistics}}}        & \mbox{~~~~}- bot utterances & 7 & 10 & 10 &  4 & 18 & 8 & 4 \\
              & \mbox{~~~~}- outputs from API calls &  0 &  0 & 23 &  7 & 24 & 40 & 0 \\
      \hline
                      & Vocabulary size       & \mcf{3,747} & 1,229 & 8,629\\
                      &Candidate set size     & \mcf{4,212} & 2,406 & 11,482\\
      {\sc Datasets} & Training dialogs       & \mcf{1,000} & 1,618 & 3,249\\
       \mco{{\small {\it Tasks 1-5 share the}}}               & Validation dialogs    & \mcf{1,000} &   500 & 403 \\
        \mco{{\small {\it same data source}}}              & Test dialogs          & \mcf{~~~~1,000$^{(*)}$} & 1,117 & 402\\
        \hline
    \end{tabular}
    }
  \end{center}
  \vspace{-3ex}
\end{table}

\subsection{Restaurant Reservation Simulation}


The simulation is based on an underlying KB, whose facts 
contain the restaurants that can be booked and their
properties. Each restaurant is defined by a type of cuisine (10
choices, e.g., French, Thai), a location (10 choices, e.g.,
London, Tokyo), a price range (cheap, moderate or expensive) and a
rating (from 1 to 8).  For simplicity, we assume that
each restaurant only has availability for a single party size (2, 4, 6
or 8 people). Each restaurant also has an address and a phone number
listed in the KB.

The KB can be queried using API calls, which return the list of facts
related to the corresponding restaurants. Each query must
contain four fields: a location, a type of cuisine, a price range and
a party size. It can return facts concerning one, several or no
restaurant (depending on the party size).

Using the KB, conversations are generated in the format shown in 
Figure~\ref{fig:tasks}.
Each example is a dialog comprising utterances from a user and a bot,
as well as API calls and the resulting facts. 
Dialogs are generated after creating a user request by sampling
an entry for each of the four required fields: e.g. the request in
Figure~\ref{fig:tasks} is [cuisine: British, location: London, party size: six,
price range: expensive].
We use natural language patterns to create user and bot
utterances. There are 43 patterns for the user and 20 for the bot (the
user can use up to 4 ways to say something, while the bot always uses
the same). Those patterns are combined with the KB entities
to form thousands of different utterances.

\subsubsection{Task Definitions}\label{sec:taskdef}

We now detail each task. Tasks 1 and 2 test dialog management to see
if end-to-end systems can learn to implicitly track dialog state
(never given explicitly), whereas Task 3 and 4 check if they can
learn to use KB facts in a dialog setting. Task 3 also requires to
learn to sort. Task 5 combines all tasks.

\vspace{-1ex}
\paragraph{Task 1: Issuing API calls}

A user request implicitly defines a query that can contain from
0 to 4 of the required fields (sampled uniformly; in
Figure~\ref{fig:tasks}, it contains 3).
The bot must ask questions for filling the missing fields and
eventually generate the correct corresponding API call.
The bot asks for information in a deterministic order, making prediction
possible.

\vspace{-1ex}
\paragraph{Task 2: Updating API calls}

Starting by issuing an API call as in Task 1,
users then ask to update their requests between 1 and 4 times (sampled
uniformly). The order in which fields are updated is random. The
bot must ask users if they are done with their updates and issue the
updated API call.

\vspace{-1ex}
\paragraph{Task 3: Displaying options}
Given a user request, we query the KB using the corresponding API call
and add the facts resulting from the call to the dialog history. 
%
The bot must propose options to users by listing the restaurant names
sorted by their corresponding rating (from higher to lower) until
users accept.
For each option, users have a 25\% chance of accepting.
If they do, the bot must stop displaying options, otherwise propose the next one.
Users always accept the option if this is the last remaining one.
We only keep examples with API calls retrieving at least 3 options.

\vspace{-1ex}
\paragraph{Task 4: Providing extra information}
Given a user request, we sample a restaurant and start the dialog as
if users had agreed to book a table there. We add all KB facts
corresponding to it to the dialog. Users then ask for the phone number
of the restaurant, its address or both, with proportions 25\%, 25\%
and 50\% respectively. The bot must learn to use the KB facts
correctly to answer.

\vspace{-1ex}
\paragraph{Task 5: Conducting full dialogs}

We combine Tasks 1-4 to generate full dialogs just as in
Figure~\ref{fig:tasks}. Unlike in Task 3,
we keep examples if API calls return at least 1 option instead of 3.

\subsubsection{Datasets}

We want to test how well models handle entities appearing
in the KB but not in the dialog training sets.
We split types of cuisine and locations in half, and create
two KBs, one with all facts about restaurants within the first halves
and one with the rest.
This yields two KBs of 4,200 facts and 600 restaurants each
(5 types of cuisine $\times$ 5 locations $\times$ 3 price ranges
$\times$ 8 ratings) that only share price ranges, ratings and party
sizes, but have disjoint sets of restaurants, locations, types of
cuisine, phones and addresses.
We use one of the KBs to generate the standard training, validation and
test dialogs, and use the other KB only to generate test dialogs, termed
Out-Of-Vocabulary (OOV) test sets.

For training, systems have access to the training examples and both
KBs. We then evaluate on both test sets, plain and OOV.
Beyond the intrinsic difficulty of each task, the challenge on
the OOV test sets is for models to generalize to new
entities (restaurants, locations and cuisine types) unseen in any
training dialog -- something natively impossible for embedding
methods. Ideally, models could, for instance, leverage information coming from
the entities of the same type seen during training.

We generate five datasets, one per task defined
in~\ref{sec:taskdef}.
Table~\ref{tab:data} gives their statistics.
Training sets are relatively small (1,000 examples) to create
realistic learning conditions.
The dialogs from the training and test sets are
different, never being based on the same user requests. Thus,
we test if models can generalize to new combinations of fields.
Dialog systems are evaluated in a ranking, not a generation, setting:
at each turn of the dialog, we test whether they can predict bot
utterances and API calls by selecting a candidate, not by generating
it.\footnote{ \cite{lowe2016evaluation} termed this setting Next-Utterance-Classification.}
Candidates are ranked from a set of all bot utterances and API calls
appearing in training, validation and test sets (plain and OOV) for
all tasks combined.
%

\subsection{Dialog State Tracking Challenge}

Since our tasks rely on synthetically generated language for
the user, we supplement our dataset with real human-bot dialogs.
We use data from DSTC2 
\citep{henderson2014second}, that is also in the restaurant booking
domain.
Unlike our tasks, its user requests only require 3 fields: type of
cuisine (91 choices), location (5 choices) and price range (3 choices).
The dataset was originally designed for dialog state tracking hence
every dialog turn is labeled with a state (a user intent + slots) to
be predicted. As our goal is to evaluate end-to-end training,
we did not use that, but instead converted the data into
the format of our 5 tasks and included it in the dataset as Task 6.

We used the provided speech transcriptions to create the user and bot
utterances, and given the dialog states we created the API calls to
the KB and their outputs which we added to the dialogs.  We also added
ratings to the restaurants returned by the API calls, so that the
options proposed by the bots can be consistently predicted (by using the
highest rating).
We did use the original test set but use a slightly different
training/validation split.
Our evaluation differs from the challenge (we do not predict
the dialog state), so we cannot compare with the results from
\citep{henderson2014second}.

%
%

This dataset has similar statistics to our Task 5 (see
Table~\ref{tab:data}) but is harder. The dialogs are noisier and
the bots made mistakes due to speech recognition errors or
misinterpretations and also do not always have a deterministic behavior
(the order in which they can ask for information varies).

\subsection{Online Concierge Service}

Tasks 1-6 are, at least partially, artificial. This provides perfect
control over their design (at least for Tasks 1-5), but no
guarantee that good performance would carry over from
such synthetic to more realistic conditions.
To quantify this, we also evaluate the models from
Section~\ref{sec:models} on data extracted from a real online
concierge service performing restaurant booking: users make requests
through a text-based chat interface that are handled by human
operators who can make API calls.
All conversations are between native English
speakers.

We collected around 4k chats to create this extra dataset, denoted
\emph{\Co}. All conversations have been anonymized by (1) removing
all user identifiers, (2) using the Stanford NER tagger to remove
named entities (locations, timestamps, etc.), (3) running some
manually defined regex to filter out any remaining salient information
(phone numbers, etc.).
The dataset does not contain results from API calls, but still records
when operators made use of an external service (Yelp or OpenTable) to
gather information. Hence, these have to be predicted, but without any
argument (unlike in Task 2).

The statistics of {\emph{\Co}} are given in
Table~\ref{tab:data}.
The dialogs are shorter than in Tasks 1-6, especially since they do
not include results of API calls, but the vocabulary is more diverse
and so is the candidate set; the candidate set is made of all
utterances of the operator appearing in the training, validation and test
sets.
Beyond the higher variability of the language used by human operators
compared to bots, the dataset offers additional challenges.
The set of user requests is much wider,
ranging from managing restaurant reservations to asking for recommendations or
specific information. Users do not always stay focused on the request. API
calls are not always used (e.g., the operator might use neither Yelp nor
OpenTable to find a restaurant), and facts about restaurants are not structured
nor constrained as in a KB. The structure of dialogs is thus much more
variable. Users and operators also make typos, spelling and grammar mistakes.










\section{Models} \label{sec:models}
\vspace{-0.5ex}
To demonstrate how to use the dataset and provide baselines,
we evaluate several learning methods on our goal-oriented dialog tasks:
rule-based systems, classical information retrieval methods, supervised
embeddings, and end-to-end Memory networks.

\subsection{Rule-based Systems}
\vspace{-0.5ex}

Our tasks T1-T5 are built with a simulator so as to be completely predictable.
Thus it is possible to hand-code a rule based system that achieves 100\%
on them, similar to the bAbI tasks of 
\cite{weston2015towards}. Indeed, the point of these tasks is not to check
whether a human is smart enough to be able to build a rule-based system to
solve them, but to help analyze in which circumstances
machine learning algorithms are smart enough to work, and where they fail.

However, the Dialog State Tracking Challenge
 task (T6) contains some real interactions with users. This makes rule-based systems less straightforward
and not so accurate (which is where we expect machine learning to be useful).
We implemented a rule-based system for this task in the following way.  We
initialized a dialog state using the 3 relevant slots for this task: cuisine
type, location and price range.  Then we analyzed the training data and wrote a
series of rules that fire for triggers like word matches, positions in the
dialog, entity detections or dialog state, to output particular responses, API
calls and/or update a dialog state. Responses are created by combining patterns
extracted from the training set with entities detected in the previous turns or
stored in the dialog state.  Overall we built 28 rules and extracted 21
patterns.  We optimized the choice of rules and their application priority
(when needed) using the validation set, reaching a validation per-response
accuracy of 40.7\%.
We did not build a rule-based system for \emph{\Co} data as it is even less constrained.

\subsection{Classical Information Retrieval Models}
\vspace{-0.5ex}

Classical information retrieval (IR) models with no machine learning are standard baselines that often perform surprisingly well on dialog tasks
\citep{isbell2000cobot,jafarpour2010filter,ritter2011data,sordoni2015neural}.
We tried two standard variants:

\vspace{-0.5ex}
\paragraph{TF-IDF Match} For each possible candidate response, we compute a matching score between the input and the response, and rank the responses by score. The score is the TF–IDF weighted cosine similarity between the bag-of-words of the input and bag-of-words of the candidate response. We consider the case of the input being either only the last utterance or the entire conversation history, and choose the variant that works best on the validation set (typically the latter).

\vspace{-0.5ex}
\paragraph{Nearest Neighbor} Using the input, we find the most similar conversation in the training set, and output the response from that example. In this case we consider the input to only be the last utterance, and consider the training set as  (utterance, response) pairs that we select from. We  use word overlap as the scoring method. When several responses are associated with the same utterance in training, we sort them by decreasing co-occurence frequency. 

\subsection{Supervised Embedding Models}\label{sem}
\vspace{-0.5ex}

A standard, often strong, baseline is to use supervised
word embedding models for scoring (conversation history, response)
pairs. The embedding vectors are trained directly for this goal.  In
contrast, word embeddings are most well-known in the context of
unsupervised training on raw text as in {\it word2vec} 
\citep{mikolov2013efficient}. Such models are trained by learning to
predict the middle word given the surrounding window of words, or
vice-versa.  However, given training data consisting of dialogs, a
much more direct and strongly performing training procedure can be
used: predict the next response given the previous
conversation.  In this setting a candidate reponse $y$ is scored
against the input $x$: $f(x,y) = (Ax)^\top By$, where $A$ and $B$ are
$d \times V$ word embedding matrices, i.e. input and response are
treated as summed bags-of-embeddings.  We also consider the case of
enforcing $A=B$, which sometimes works better, and optimize the choice
on the validation set.

The embeddings are trained with a margin ranking loss:
$f(x,y) > m + f(x,\bar{y})$, with $m$ the size of the margin,  
and we sample $N$ {\em negative} candidate responses $\bar{y}$ per example, and train with SGD.
This approach has been previously shown to be very effective in a range of contexts
\citep{bai2009supervised,dodge2015evaluating}.
This method can be thought of as a classical information retrieval model, but where the matching function is learnt.

\subsection{Memory Networks}\label{sec:memnn}
\vspace{-0.5ex}

Memory Networks \citep{weston2014memory,sukhbaatar2015end} are a
recent class of models that have been applied to a range of natural
language processing tasks, including question answering
\citep{weston2015towards}, language modeling
\citep{sukhbaatar2015end}, and non-goal-oriented
dialog \citep{dodge2015evaluating}.  By first writing and then
iteratively reading from a memory component (using {\it hops}) that can store historical
dialogs and short-term context to reason about the required response,
they have been shown to perform well on those tasks and to outperform some other
end-to-end architectures based on Recurrent Neural Networks. Hence, we chose them as end-to-end model baseline. 

We use the MemN2N architecture of \citet{sukhbaatar2015end}, with an
additional modification to leverage exact matches and types, described shortly.
Apart from that addition, the main components of the model are (i) how it
stores the conversation in memory, (ii) how it reads from the memory 
to reason about the response; and (iii) how it outputs the response.  The
details are given in Appendix~\ref{sec:memnn}.

\vspace{-0.5ex}
\subsection{Match type features to deal with entities}
Words denoting entities have two important traits:
1) exact matches are usually more appropriate to deal with them
than approximate matches, and 2) they frequently appear as OOV words
(e.g., the name of a new restaurant).
Both are a challenge for embedding-based methods.
Firstly, embedding into a low dimensional space makes it hard to
differentiate between {\em exact} word matches, and matches between
words with similar meaning \citep{bai2009supervised}.  While this can
be a virtue (e.g. when using synonyms), it is often a flaw
when dealing with entities
(e.g. failure to differentiate between phone numbers since they have similar embeddings).
Secondly, when a new word is used (e.g. the name of a new restaurant)
not seen before in training, no word embedding is available,
typically resulting in failure \citep{weston2014memory}.

Both problems can be alleviated with match type features.
Specifically, we augment the vocabulary with 7 special words,
one for each of the KB entity types
(cuisine type, location, price range, party size, rating, phone number and
address). For each type, the corresponding type word is added to the
candidate representation if a word is found that appears 1) as a KB entity
of that type, 2) in the candidate, and 3) in the input or memory.
Any word that matches as a KB entity can be typed
even if it has never been seen before in training dialogs.
These features allow the model to learn to rely on type information
 using exact matching words cues when OOV entity embeddings are not known, as
 long as it has access to a KB with the OOV entities.
 We assess the impact of such features for TF-IDF Match, Supervised
 Embeddings and Memory Networks.

\section{Experiments}\label{sec:exp}

\begin{table}[t!]
  \begin{center}
    \caption{{\bf Test results across all tasks and methods.} {\small
For tasks T1-T5 results are given in the standard setup and the out-of-vocabulary (OOV) setup, where words (e.g. restaurant names) may not have been seen during training. Task T6 is the Dialog state tracking 2 task with real dialogs, and only has one setup.
Best performing methods (or methods within 0.1\% of best performing) are given in bold for the per-response accuracy metric, with the per-dialog accuracy given in parenthesis.
$^{(*)}$ For \Co, an example is considered correctly answered if the correct response is ranked among the top 10 candidates by the bot,
to accommodate the much larger range of semantically equivalent responses among candidates (see ex. in Tab.~\ref{tab:concierge1})
\label{tab:res1}. $^{(\dagger)}$ We did not implement MemNNs+match type on \Co, because this method requires a KB and there is none associated with it.}
}
    \resizebox{1.01\linewidth}{!}{
      {
	      \begin{tabular}{l|c@{\,}c|c@{\,}cc@{\,}cc@{\,}cc@{\,}cc@{\,}cc@{\,}c}
	      Task & \mcb{Rule-based} & \multicolumn{4}{c}{TF-IDF Match} & \mc{Nearest}   & \mc{Supervised} & \multicolumn{4}{c}{Memory Networks}\\
					& \mcb{Systems} & \mc{no type} & \mc{+ type} & \mc{Neighbor} & \mc{Embeddings} &    \mc{no match type}        & \mc{+ match type}\\
      \hline
      T1: Issuing API calls         &\it 100 &\it (100)& 5.6 &(0)& 22.4 &(0)& 55.1 &(0)& \bf 100 &(100)& \bf 99.9 &(99.6)& \bf 100  & (100)\\
      T2: Updating API calls        &\it 100 &\it (100)& 3.4 &(0)& 16.4 &(0)& 68.3 &(0)&     68.4  &(0)& \bf 100  & (100)&     98.3 &(83.9)\\
      T3: Displaying options        &\it 100 &\it (100)& 8.0 &(0)& 8.0 &(0)& 58.8 &(0)&     64.9  &(0)& \bf 74.9 & (2.0)&  \bf 74.9 &   (0)\\
      T4: Providing information     &\it 100 &\it (100)& 9.5 &(0)& 17.8 &(0)& 28.6 &(0)&     57.2  &(0)&     59.5 & (3.0)& \bf 100  & (100)\\
      T5: Full dialogs              &\it 100 &\it (100)& 4.6 &(0)& 8.1 &(0)& 57.1 &(0)&     75.4  &(0)& \bf 96.1 &(49.4)&     93.4 &(19.7)\\
      \hline
      T1(OOV): Issuing API calls     &\it 100 &\it (100)& 5.8 &(0)&22.4 &(0)& 44.1 &(0)&     60.0  &(0)&     72.3 &   (0)& \bf 96.5 &(82.7)\\
      T2(OOV): Updating API calls    &\it 100 &\it (100)& 3.5 &(0)&16.8 &(0)& 68.3 &(0)&     68.3  &(0)&     78.9 &   (0)& \bf 94.5 &(48.4)\\
      T3(OOV): Displaying options    &\it 100 &\it (100)& 8.3 &(0)&8.3 &(0)& 58.8 &(0)&     65.0  &(0)&     74.4 &   (0)& \bf 75.2 &   (0)\\
      T4(OOV): Providing inform. &\it 100 &\it (100)& 9.8 &(0)& 17.2 &(0)& 28.6 &(0)&     57.0  &(0)&     57.6 &   (0)& \bf 100  & (100)\\
      T5(OOV): Full dialogs          &\it 100 &\it (100)& 4.6 &(0)&9.0 &(0)& 48.4 &(0)&     58.2  &(0)&     65.5 &   (0)& \bf 77.7 &   (0)\\
      \hline
      T6: Dialog state tracking 2   &    33.3&      (0)& 1.6 &(0)&1.6 &(0)& 21.9 &(0)&     22.6  &(0)& \bf 41.1 &   (0)&   \bf 41.0 &   (0)\\
      \hline     
      \hline      
      \Co$^{(*)}$  & n/a & & 1.1 & (0.2)& n/a &&13.4& (0.5)& 14.6& (0.5)& \bf 16.7& (1.2)& n/a$^{(\dagger)}$ & \\
    \end{tabular}
    }
    }
  \end{center}
\end{table}

Our main results across all the models and tasks are given in Table
\ref{tab:res1} (extra results are also given in Table~\ref{tab:res2} of Appendix~\ref{sec:more-res}).  The first 5 rows show tasks T1-T5, and rows 6-10 show
the same tasks in the out-of-vocabulary setting. Rows 11 and 12 give
results for the Dialog State Tracking Challenge task (T6) and
\Co~respectively.  Columns 2-7 give the results of each method tried
in terms of per-response accuracy and per-dialog accuracy, the latter
given in parenthesis.  Per-response accuracy counts the percentage of
responses that are correct (i.e., the correct candidate is chosen out
of all possible candidates).  Per-dialog accuracy counts the
percentage of dialogs where every response is correct. Ultimately, if
only one response is incorrect this could result in a failed dialog,
i.e. failure to achieve the goal (in this case, of achieving a
restaurant booking).  Note that we test Memory Networks (MemNNs) with
and without match type features, the results are shown in the last two
columns.  The hyperparameters for all models were optimized on the
validation sets; values for best performing models are given in
Appendix~\ref{sec:hp}.


The classical IR method TF-IDF Match performs the worst of all methods, and 
much worse than the Nearest Neighbor IR method, which is true on both
the simulated tasks  T1-T5 and on the real data of T6 and \Co.
Supplementing TF-IDF Match with match type features noticeably improves
performance, which however still remains far behind Nearest Neighbor IR
(adding bigrams to the dictionary has no effect on performance).
This is in sharp contrast to
other recent results on data-driven {\em non}-goal directed conversations, e.g. 
over dialogs on Twitter \citep{ritter2011data} or Reddit \citep{dodge2015evaluating}, where it was found that TF-IDF Match outperforms Nearest Neighbor,
as general conversations on a given subject typically share many words.
We conjecture that the goal-oriented nature of the conversation means that the 
conversation moves forward more quickly, sharing fewer words per
(input, response) pair, e.g. consider the example in Figure \ref{tab:data}.

Supervised embeddings outperform classical IR methods in general,
indicating that learning mappings between words (via word embeddings)
is important.  However, only one task (T1, Issuing API calls) is
completely successful.  In the other tasks, some responses are
correct, as shown by the per-response accuracy, however there is no
dialog where the goal is actually achieved (i.e., the mean
dialog-accuracy is 0). Typically the model can provide correct
responses for greeting messages, asking to wait, making API calls and
asking if there are any other options necessary. However, it fails to
interpret the results of API calls to display options, provide
information or update the calls with new information, resulting in
most of its errors, even when match type features are provided.
%

 Memory Networks (without match type features) outperform classical IR and
supervised embeddings across all of the tasks. They can solve the first two
tasks (issuing and updating API calls) adequately. On the other tasks,
they give improved results, but do not solve them. While the per-response
accuracy is improved, the per-dialog accuracy is still close to 0 on T3 and T4.
Some examples of predictions of the MemNN for T1-4 are given in Appendix~\ref{sec:ap}.
On the OOV tasks again performance is improved, but this is all due to
better performance on {\em known} words, as unknown words are simply
not used without the match type features. As stated in
Appendix~\ref{sec:hp}, optimal hyperparameters on several of the tasks
involve 3 or 4 hops, indicating that iterative accessing and reasoning
over the conversation helps, e.g. on T3 using 1 hop gives 64.8\% while
2 hops yields 74.7\%. Appendix~\ref{sec:ap} displays illustrative
examples of Memory Networks predictions on T 1-4 and \Co.

Memory Networks {\em with match type features} give two performance gains over the same models without match type features: (i) T4 (providing information) becomes solvable because matches can be made to the results of the API call; and (ii) 
out-of-vocabulary results are significantly improved as well. Still, tasks T3 and T5 are still fail cases, performance drops slightly on T2 compared to not using match type features, and no relative improvement is observed on T6.
Finally, note that matching words on its own is not enough,
 as evidenced by the poor performance of TF-IDF matching; this idea must be combined with types and the other properties of the MemNN model.

Unsurprisingly, perfectly coded rule-based systems can solve the 
simulated tasks T1-T5 perfectly, whereas our machine learning methods cannot.
However, it is not easy to build an effective rule-based system when
dealing with real language on real problems, and our rule based system is
outperformed by MemNNs on the more realistic task T6.


Overall, while the methods we tried made some
inroads into these tasks, there are still many challenges left unsolved.
Our best models can learn to track implicit dialog states and
manipulate OOV words and symbols (T1-T2) to issue API calls and
progress in conversations, but they are still unable to perfectly
handle interpreting knowledge about entities (from returned API calls)
to present results to the user, e.g. displaying options in T3.  The
improvement observed on the simulated tasks e.g. where MemNNs
outperform supervised embeddings which in turn outperform IR methods,
is also seen on the realistic data of T6 with similar relative
gains. This is encouraging as it indicates that future work on
breaking down, analysing and developing models over the simulated
tasks should help in the real tasks as well.
Results on \Co~confirm this observation: the pattern of
relative performances of methods is the same
on \Co~and on our series of tasks. This suggests that our synthetic data can
indeed be used as an effective evaluation proxy.

\section{Conclusion}
We have introduced an open dataset and task set for evaluating end-to-end
goal-oriented dialog learning methods in a systematic and controlled way. 
We hope this will help foster progress of end-to-end conversational
agents because (i) existing measures of performance either prevent reproducibility (different Mechanical Turk jobs) or do not correlate well with human judgements \citep{liu2016not}; (ii) the breakdown in tasks will
 help focus research and development to improve the learning methods;
 and (iii) goal-oriented dialog has clear utility in real applications.
We illustrated how to use the testbed using a variant of end-to-end Memory
Networks, which prove an effective model on these tasks relative to other
baselines, but are still lacking in some key areas. 



\subsection*{Acknowledgments}
The authors would like to thank Martin Raison, Alex Lebrun and Laurent Landowski for their help with the \Co~ data.

\small
\bibliography{dialog}
\bibliographystyle{natbib}

\appendix
\section{Memory Networks implementation} \label{sec:memnn}
\paragraph{Storing and representing the conversation history}
As the model conducts a conversation with the user, at each time step
$t$ the previous utterance (from the user) and response (from the
model) are appended to the memory.  Hence, at any given time there are
$c^u_1, \dots c^u_{t}$ user utterances and $c^r_1, \dots c^r_{t-1}$
model responses stored (i.e. the entire conversation).\footnote{API calls are stored as bot utterances $c^r_i$, and KB facts resulting from such calls as user utterances $c^u_i$.
} The aim at time
$t$ is to thus choose the next response $c^r_{t}$.  We train on
existing full dialog transcripts, so at training time we know the
upcoming utterance $c^r_{t}$ and can use it as a training target.
Following \cite{dodge2015evaluating}, we represent each utterance as a
bag-of-words and in memory it is represented as a vector using the
embedding matrix $A$, i.e. the memory is an array with entries:
\[
    m = (A \Phi(c^u_1), A \Phi(c^r_1) \dots, A \Phi(c^u_{t-1}), A \Phi(c^r_{t-1}))
\]
where $\Phi(\cdot)$ maps the utterance to a bag of dimension $V$ (the
vocabulary), and $A$ is a $d \times V$ matrix, where $d$ is the
embedding dimension.  We retain the last user utterance $c^u_t$ as the
``input'' to be used directly in the controller.  The contents of each
memory slot $m_i$ so far does not contain any information of which
speaker spoke an utterance, and at what time during the conversation.
We therefore encode both of those pieces of information in the mapping
$\Phi$ by extending the vocabulary to contain
$T=1000$ extra ``time features'' which encode the index $i$ into the
bag-of-words, and two more features that encode whether the utterance
was spoken by the user or the model.

\vspace{-0.5ex}
\paragraph{Attention over the memory} 
The last user utterance $c^u_{t}$ is embedded using the same matrix $A$ giving $q=A \Phi(c^u_{t})$, which can
also be seen as the initial state of the controller. At this point the controller reads from the memory to find
salient parts of the previous conversation that are relevant to producing a response.
The match between $q$ and the memories is  computed by taking the inner product followed by a softmax: 
$p_i = \text{Softmax}(u^\top m_i)$,
giving a probability vector over the memories.
The vector that is returned back to the controller 
is then computed by $o = R \sum_i p_i m_i$ where $R$ is a $d \times d$
square matrix.
The controller state is then updated with $q_2 = o + q$.
The memory can be iteratively reread to look for additional pertinent information using the updated
state of the controller $q_2$ instead of $q$, and in general using $q_h$ on 
iteration $h$, with a fixed number of iterations $N$ (termed $N$ hops).
Empirically we find improved performance on our tasks with up to 3 or 4 hops.

\vspace{-0.5ex}
\paragraph{Choosing the response} The final prediction is then defined as:
\[
  \hat{a} = \text{Softmax}( {q_{N+1}}^{\top} W \Phi(y_1), \dots, {q_{N+1}}^{\top} W \Phi(y_C))
\]
 where there are $C$ candidate responses in $y$,
and $W$ is of dimension $d \times V$. In our tasks the set $y$ is a (large) set of candidate responses which includes all possible bot utterances and API calls.

The entire model is trained using stochastic gradient descent (SGD),
minimizing a standard cross-entropy loss between $\hat{a}$ and the true label $a$.

\section{Examples of Predictions of a Memory Network} \label{sec:ap}
\newcommand{\mcq}[1]{\multicolumn{2}{|l|}{{\bf #1}}}
\newcommand{\mct}[1]{\multicolumn{1}{l}{{ #1}}}
\newcommand{\mce}[1]{\multicolumn{3}{l|}{{\bf #1}}}
\newcommand{\mcee}[1]{\multicolumn{2}{l|}{{\bf #1}}}
\newcommand{\mces}[1]{\multicolumn{2}{l|}{{#1}}}

Tables \ref{tab:ext1}, \ref{tab:ext2}, \ref{tab:ext3} and
\ref{tab:ext4} display examples of predictions of the best
performing Memory Network on full dialogs, Task 5, (with 3 hops) on test examples of Tasks 1-4
along with the values of the attention over each memory for each hop ($p_i$ as
defined in Sec.~\ref{sec:memnn}). This model does not use match type features.
Then, Table \ref{tab:concierge1} displays an example of prediction of
the best performing Memory Network on \Co~(with 2 hops) on a test
example along with the values of the attention over each memory for
each hop.

\begin{table}[htbp!]
  \begin{center}
   \caption{\label{tab:ext1} {\bf Task 1 (Issue API call)} {\small The model
     learns to direct its attention towards the 4 memories containing
     the information key to issue the API call. More hops help to
     strengthen this signal. <silence> is a special token used to
     indicate that the user did not speak at this turn -- the model
     has to carry out the conversation with no additional input.}}
   {\small
  \begin{tabular}{|l|l|l|c|c|c|}
    \hline
Time & Locutor & Dialog History & Hop \#1 & Hop \#2 & Hop \#3 \\
    \hline
1 & User&  hi &                                  .054  &  .016 &   .020\\
2 & Bot&  hello what can i help you with today &  .040  &  .024  &  .008 \\
3 & User&  may i have a table in paris &            .099 & \cellcolor[rgb]{1,0.8,0.8}.137 &\cellcolor[rgb]{1,0.8,0.8}.141 \\
4 & Bot&  i'm on it &                            .048  &  .028  &  .004 \\
5 & User&  <silence> &                            \cellcolor[rgb]{1,0.8,0.8}.181 &   .099  &  .037 \\
6 & Bot&  any preference on a type of cuisine &     .056 &   .090 &   .014 \\
7 & User&  i love indian food &                   \cellcolor[rgb]{1,0.8,0.8}.159 & \cellcolor[rgb]{1,0.8,0.8}.188  & \cellcolor[rgb]{1,0.5,0.5}.238 \\
8 & Bot& how many people would be in your party &  .051 &   .022  &  .010 \\
9 & User&  we will be six &                        \cellcolor[rgb]{1,0.8,0.8}.129 & \cellcolor[rgb]{1,0.8,0.8}.144  & \cellcolor[rgb]{1,0.8,0.8}.114 \\
10 & Bot&  which price range are looking for &      .039 &   .028  &  .006 \\
11 & User&  in a moderate price range please &   \cellcolor[rgb]{1,0.8,0.8}.128 &  \cellcolor[rgb]{1,0.5,0.5}.213  & \cellcolor[rgb]{1,0,0}.405 \\
12 & Bot&  ok let me look into some options for you &       .016 &   .011  &  .003 \\
\hline
\mcq{User input} &  \mct{<silence>} & \mce{ } \\
\hline
\mcq{Correct answer} & \mct{{\tt api\_call indian paris six moderate}}  & \mce{ } \\
\hline
\mcq{Predicted answer} & \mct{{\tt api\_call indian paris six moderate}}  & \mce{[Correct] } \\
\hline
   \end{tabular}}
   \end{center}
  \end{table}

\begin{table}[htbp!]
  \begin{center}
  \caption{\label{tab:ext2} {\bf Task 2 (Update API call)} {\small Out of the
    multiple memories from the current dialog, the model correctly
    focuses on the 2 important pieces: the original API call and the
    utterance giving the update.}}
  {\small
  \begin{tabular}{|l|l|l|c|c|c|}
    \hline
Time & Locutor & Dialog history & Hop \#1 & Hop \#2 & Hop \#3 \\
\hline
3 & User&  may i have a table in paris       &   .061 &   .072 &   .040\\
4 & Bot& i'm on it                          &  .026  &  .012  &  .001\\
5 & User& <silence>                          &  .087 &   .042  &  .012\\
6 & Bot&  any preference on a type of cuisine& .026  &  .023  &  .001\\
7 & User& i love indian food                 & .081  &  .070  &  .055\\
8 & Bot& how many people would be in your party&       .025  &  .006  &  .001\\
9 & User& we will be six                       & .059  &  .051  &  .018\\
10 & Bot& which price range are looking for    &  .038 &   .043 &   .004\\
11 & User& in a moderate price range please    &  .080  &  .095  &  .096\\
12 & Bot& ok let me look into some options for you &  .025  &  .042 &   .003\\
13 & User& <silence>                            &  \cellcolor[rgb]{1,0.8,0.8}.127  &  .069  &  .032\\
14 & Bot& \tt api\_call indian paris six moderate   &  .062  &  \cellcolor[rgb]{1,0.8,0.8}.113  &  .043\\
15 & User& instead could it be with french food  &  \cellcolor[rgb]{1,0.8,0.8} .188  & \cellcolor[rgb]{1,0.5,0.5}.311 & \cellcolor[rgb]{1,0,0}.683\\
16 & Bot& sure is there anything else to update  &  .016  &  .007 &   .001\\
17 & User& no & .028  &  .013  &  .007\\
18 & Bot& ok let me look into some options for you  &   .011  &  .006 &   .000\\
\hline
\mcq{User input} &  \mct{<silence>} & \mce{ } \\
\hline
\mcq{Correct answer} & \mct{{\tt api\_call french paris six moderate}}  & \mce{ } \\
\hline
\mcq{Predicted answer} & \mct{{\tt api\_call french paris six moderate}}  & \mce{[Correct] } \\
\hline
  \end{tabular}}
  \end{center}
  \end{table}

\begin{table}[htbp!]
  \begin{center}
    \caption{\label{tab:ext3} {\bf Task 3 (Displaying options)} {\small The
      model knows it has to display options but the attention is
      wrong: it should attend on the ratings to select the best
      option (with highest rating). It cannot learn that
      properly and match type features do not help. It is correct here by luck,
the task is not solved overall (see Tab.~\ref{tab:res1}). We do not show all
memories in the table, only those with meaningful attention.}}
    {\small
  \begin{tabular}{|l|l|l|c|c|c|}
    \hline
Time & Locutor & Dialog history & Hop \#1 & Hop \#2 & Hop \#3 \\
\hline
14 & Bot& \tt api\_call indian paris six moderate & .012   & .000   & .000 \\
15 & User& instead could it be with french food & .067   & \cellcolor[rgb]{1,0.8,0.8}.103   &  \cellcolor[rgb]{1,0.8,0.8}.147 \\
20 & Bot& \tt api\_call french paris six moderate  & .012   & .000   & .000 \\
21 & User& \tt resto\_1 r\_phone rest\_1\_phone & .018   & .004   & .000 \\
23 & User& \tt resto\_1 r\_cuisine french & .029   & .005   & .000 \\
24 & User& \tt resto\_1 r\_location paris  & .060   & \cellcolor[rgb]{1,0.5,0.5}.292   & .094 \\
25 & User& \tt resto\_1 r\_number six & .050   &\cellcolor[rgb]{1,0.5,0.5}.298   & \cellcolor[rgb]{1,0,0}.745 \\
26 & User& \tt resto\_1 r\_price moderate & .060   & .090   & .002 \\
27 & User& \tt resto\_1 r\_rating 6& .016   & .002   & .000 \\
30 & User& \tt resto\_2 r\_cuisine french  & .031   & .007   & .000 \\
31 & User& \tt resto\_2 r\_location paris & .040   & .081   & .004 \\
32 & User& \tt resto\_2 r\_number six & .020   & .012   & .000 \\
33 & User& \tt resto\_2 r\_price moderate & .029   & .009   & .000 \\
37 & User& \tt resto\_3 r\_cuisine french & .014   & .001   & .000 \\
38 & User& \tt resto\_3 r\_location paris & .028   & .016   & .001 \\
39 & User& \tt resto\_3 r\_number six & .024   & .022   & .004 \\
40 & User& \tt resto\_3 r\_price moderate & .039   & .015   & .001 \\
\hline
\mcq{User input} &  \mct{<silence>} & \mce{ } \\
\hline
\mcq{Correct answer} & \mct{what do you think of this option: resto\_1}  & \mce{ } \\
\hline
\mcq{Predicted answer} & \mct{what do you think of this option: resto\_1}  & \mce{[Correct] } \\
\hline
    \end{tabular}}
    \end{center}
  \vspace*{5ex}
\end{table}

\begin{table}[htbp!]
  \begin{center}
  \caption{\label{tab:ext4} {\bf Task 4 (Providing extra-information)}
    {\small The model knows it must display a phone or an address, but, as
    explained in Section~\ref{sec:memnn} the embeddings mix up the
    information and make it hard to distinguish between different
    phone numbers or addresses, making answering correctly very
    hard. As shown in the results of Tab.~\ref{tab:res1}, this problem can be solved
    by adding match type features, that allow to emphasize entities
    actually appearing in the history. The attention is globally wrong
    here.}}  {\small
  \begin{tabular}{|l|l|l|c|c|c|}
    \hline
Time & Locutor & Dialog history & Hop \#1 & Hop \#2 & Hop \#3 \\
\hline
14 & Bot& \tt api\_call indian paris six moderate  & .006   & .000   & .000 \\
15 & User& instead could it be with french food  & .024   & .011   & .007 \\
20 & Bot&  \tt api\_call french paris six moderate    & .005   & .000   & .001 \\
21 & User& \tt resto\_1 r\_phone resto\_1\_phone  & .011   & .005   & .004 \\
22 & User& \tt resto\_1 r\_address resto\_1\_address   & .018   & .004   & .001 \\
23 & User& \tt resto\_1 r\_cuisine french         & .018   & .003   & .001 \\
24 & User& \tt resto\_1 r\_location paris    & .068   & .091   & \cellcolor[rgb]{1,0.8,0.8}.108 \\
25 & User& \tt resto\_1 r\_number six     & .086   & .078   & .020 \\
26 & User& \tt resto\_1 r\_price moderate           & .070   & \cellcolor[rgb]{1,0.5,0.5}.225   & \cellcolor[rgb]{1,0,0}.369 \\
27 & User& \tt resto\_1 r\_rating 6      & .014   & .006   & .008 \\
28 & User& \tt resto\_2 r\_phone resto\_2\_phone  & .015   & .009   & .006 \\
29 & User& \tt resto\_2 r\_address resto\_2\_address  & .014   & .004   & .001 \\
31 & User& \tt resto\_2 r\_location paris    & .075   & \cellcolor[rgb]{1,0.8,0.8}.176   & \cellcolor[rgb]{1,0.5,0.5}.193 \\
32 & User& \tt resto\_2 r\_number six     & \cellcolor[rgb]{1,0.8,0.8}.100   & \cellcolor[rgb]{1,0.8,0.8}.126   & .026 \\
33 & User& \tt resto\_2 r\_price moderate              & .038   & .090   & \cellcolor[rgb]{1,0.8,0.8}.167 \\
35 & User& \tt resto\_3 r\_phone resto\_3\_phone & .004   & .001   & .001 \\
36 & User& \tt resto\_3 r\_address resto\_3\_address  & .005   & .002   & .001 \\
37 & User& \tt resto\_3 r\_location paris            & .028   & .028   & .026 \\
39 & User& \tt resto\_3 r\_number six     & .039   & .013   & .002 \\
40 & User& \tt resto\_3 r\_price moderate  & .018   & .008   & .013 \\
42 & Bot& what do you think of this option: resto\_1   & .074   & .001   & .000 \\
43 & User& let's do it   & .032   & .004   & .001 \\
44 & Bot& great let me do the reservation & .003   & .000   & .000 \\
\hline
\mcq{User input} &  \mct{do you have its address} & \mce{ } \\
\hline
\mcq{Correct answer} & \mct{here it is resto\_1\_address}  & \mce{ } \\
\hline
\mcq{Predicted answer} & \mct{here it is:  resto\_8\_address}  & \mce{[Incorrect] } \\
\hline
  \end{tabular}}
  \end{center}
\end{table}

\begin{table}[h!]
  \begin{center}
    \caption{\label{tab:concierge1} {\bf Concierge Data} {\small The model is
      also able to learn from human-human dialogs.  <person>, <org>,
      <number> and <date> are special tokens used to anonymize the
      data. We report the top 5 answers predicted by the model. They
      are all semantically equivalent. Note that the utterances, while
      all produced by humans, are not perfect English ("rservation",
      "I'll check into it")
    }}
            {\small
  \begin{tabular}{|l|l|l|c|c|}
    \hline
Time & Locutor & Dialog History & Hop \#1 & Hop \#2 \\
    \hline
1 & User&  hey concierge &    \cellcolor[rgb]{1,0.8,0.8}                              .189  &  .095 \\
2 & User&  could you check if i can get a rservation at <org> <date> for brunch &\cellcolor[rgb]{1,0.5,0.5}.209 & \cellcolor[rgb]{1,0.8,0.8}.178 \\
3 & User&  <number> people &                            \cellcolor[rgb]{1,0.8,0.8}.197 & \cellcolor[rgb]{1,0.8,0.8}  .142  \\
4 & User&  <silence> &                            \cellcolor[rgb]{1,0.8,0.8}.187 &  \cellcolor[rgb]{1,0.8,0.8} .167  \\
5 & Bot& hi <person> unfortunately <org> is fully booked for <date> &\cellcolor[rgb]{1,0.5,0.5}     .225 & \cellcolor[rgb]{1,0,0}  .410 \\
& & and there's <number> people on the waiting list & & \\
\hline
\mcq{User input} &  \mct{when's the earliest availability} & \mcee{ } \\
\hline
\mcq{Correct answer} & \mct{i'll check}  & \mcee{ } \\
\hline
\mcq{Pred. answer \#1} & \mct{i'm on it}  & \mces{[Incorrect]} \\
\mcq{Pred. answer \#2} & \mct{i'll find out}  & \mces{[Incorrect]} \\
\mcq{Pred. answer \#3} & \mct{i'll take a look}  & \mces{[Incorrect]} \\
\mcq{Pred. answer \#4} & \mct{i'll check}  & \mcee{[Correct]} \\
\mcq{Pred. answer \#5} & \mct{i'll check into it}  & \mces{[Incorrect]} \\
\hline
   \end{tabular}}
   \end{center}
  \end{table}

\section{Hyperparameters} \label{sec:hp}

Tables~\ref{tab:hp_supemb} and~\ref{tab:hp_memnet} respectively
display the values of the hyperparameters of the best Supervised
Embeddings and Memory Networks selected for each task. These models
were selected using the best validation validation sets.

\begin{table}[htbp!]
  \begin{center}
    \caption{\label{tab:hp_supemb} {\bf Hyperparameters of Supervised
        Embeddings.} {\small When Use History is True, the whole conversation
      history is concatenated with the latest user utterance to create
      the input. If False, only the latest utterance is used as
      input.}}
      \begin{tabular}{|l|c|c|c|c|c|c|}
       \hline
       Task & Learning Rate & Margin $m$ & Embedding Dim $d$  & Negative Cand. $N$ & Use History\\
       \hline
       Task 1 & 0.01 & 0.01 & 32 & 100 & True\\
       Task 2 & 0.01 & 0.01 & 128 & 100 & False \\
       Task 3 & 0.01 & 0.1 & 128 & 1000 & False \\
       Task 4 & 0.001 & 0.1 & 128 & 1000 & False \\
       Task 5 & 0.01 & 0.01 & 32 & 100 & True \\
       \hline
       Task 6 & 0.001 & 0.01 & 128 & 100 & False\\
       \hline
       \Co & 0.001 & 0.1 & 64 & 100 & False\\
       \hline
      \end{tabular}
   \end{center}
  \end{table}

\begin{table}[htbp!]
  \begin{center}
    \caption{\label{tab:hp_memnet} {\bf Hyperparameters of Memory
        Networks.} {\small The longer and more complex the dialogs are, the more hops are needed.}}
      \begin{tabular}{|l|c|c|c|c|c|}
       \hline
       Task & Learning Rate & Margin $m$ & Embedding Dim $d$  & Negative Cand. $N$ & Nb Hops \\
       \hline
       Task 1 & 0.01 & 0.1 & 128 & 100 & 1\\
       Task 2 & 0.01 & 0.1 & 32 & 100 & 1 \\
       Task 3 & 0.01 & 0.1 & 32 & 100 & 3 \\
       Task 4 & 0.01 & 0.1 & 128 & 100 & 2 \\
       Task 5 & 0.01 & 0.1 & 32 & 100 & 3 \\
       \hline
       Task 6 & 0.01 & 0.1 & 128 & 100 & 4 \\
       \hline
       \Co & 0.001 & 0.1 & 128 & 100 & 2\\
       \hline
      \end{tabular}
   \end{center}
  \end{table}

\section{Additional Results} \label{sec:more-res}
Table~\ref{tab:res2} provides results for additional variants of supervised
embeddings, using either a dictionary that includes all bigrams to leverage some
word order information, or match type features.
On some tasks, supervised embeddings perform better when the last user
utterance is used as sole input, without the full dialog history (see
Table~\ref{tab:hp_supemb}). When no history is used, we slightly adapt match
type features to only record type: a special word corresponding to type $T$ (e.g.,
phone, address, etc) is appended to the representation of a candidate if the
candidate contains a word that appears in the knowledge base as an entity of
type $T$, regardless of whether the same word appeared earlier in the conversation.
As seen on Table~\ref{tab:res2}, match type features improve performance on out-of-vocabulary
tasks 1 and 5, bringing it closer to that of Memory Networks without match type features, but
still quite lagging Memory Networks with match type features. Bigrams slightly hurt rather
than help performance, except in Task 5 in the standard in-vocabulary setup (performance
is lower in the OOV setup).

\begin{table}[h!]
  \begin{center}
    \caption{{\bf Test results across all tasks and methods.} {\small
For tasks T1-T5 results are given in the standard setup and the out-of-vocabulary (OOV) setup, where words (e.g. restaurant names) may not have been seen during training. Task T6 is the Dialog state tracking 2 task with real dialogs, and only has one setup.
Best performing methods (or methods within 0.1\% of best performing) are given in bold for the per-response accuracy metric, with the per-dialog accuracy given in parenthesis.
}
\label{tab:res2}}
    \resizebox{1.01\linewidth}{!}{
      {
	      \begin{tabular}{l|c@{\,}cc@{\,}cc@{\,}c|c@{\,}cc@{\,}c}
	         &  \multicolumn{6}{c|}{Supervised Embeddings} & \multicolumn{4}{c}{Memory Networks}\\
	      Task &  \mc{no match type} & \mc{+ match type} & \multicolumn{2}{c|}{+ bigrams} & \mc{no match type}        & \mc{+ match type}\\
				 &  \mc{no bigram} &  \mc{no bigram}& \multicolumn{2}{c|}{no match type}& & & &  \\
      \hline
      T1: Issuing API calls         & \bf 100 &(100)& 83.2 &(0)& 98.6 &(92.4)&  \bf 99.9 &(99.6)& \bf 100  & (100)\\
      T2: Updating API calls        &    68.4 &(0)  & 68.4 &(0) & 68.3 &(0) &  \bf 100  & (100)&     98.3 &(83.9)\\
      T3: Displaying options        &    64.9 &(0)  & 64.9 &(0) & 64.9 &(0) &  \bf 74.9 & (2.0)& \bf 74.9 &   (0)\\
      T4: Providing information     &    57.2 &(0)  & 57.2 &(0) & 57.3 &(0) &      59.5 & (3.0)& \bf 100  & (100)\\
      T5: Full dialogs              &    75.4 &(0)  & 76.2 &(0) & 83.4 &(0) &  \bf 96.1 &(49.4)&     93.4 &(19.7)\\
      \hline
      T1(OOV): Issuing API calls    & 60.0  &(0) & 67.2 &(0)& 58.8 &(0) &    72.3 &   (0)& \bf 96.5 &(82.7)\\
      T2(OOV): Updating API calls   & 68.3  &(0) & 68.3 &(0)& 68.3 &(0) &    78.9 &   (0)& \bf 94.5 &(48.4)\\
      T3(OOV): Displaying options   & 65.0  &(0) & 65.0 &(0)& 62.1 &(0) &    74.4 &   (0)& \bf 75.2 &   (0)\\
      T4(OOV): Providing inform.    & 57.0  &(0) & 57.1 &(0)& 57.0 &(0) &    57.6 &   (0)& \bf 100  & (100)\\
      T5(OOV): Full dialogs         & 58.2  &(0) & 64.4 &(0)& 50.4 &(0) &    65.5 &   (0)& \bf 77.7 &   (0)\\
      \hline
      T6: Dialog state tracking 2   & 22.6  &(0) &22.1  &(0)& 21.8 &(0) & \bf 41.1 &  (0)& \bf 41.0 &   (0)\\
      \hline     
    \end{tabular}
    }
    }
  \end{center}
\end{table}

\end{document}